\documentclass[letterpaper, 10 pt, conference]{ieeeconf}  %

\IEEEoverridecommandlockouts                              %

\usepackage{graphicx} %
\usepackage{svg}
\usepackage{blindtext}

\usepackage{amsmath} %
\usepackage{amssymb}  %
\usepackage{amsthm}
\usepackage{}
\newcommand{\mycomment}[1]{}

\newcommand{\fref}[1]{Fig.~\ref{#1}}
\newcommand{\tref}[1]{Table~\ref{#1}}
\newcommand{\sref}[1]{Sec.~\ref{#1}}

\usepackage{paralist}
\usepackage{microtype}
\usepackage{multirow}
\usepackage{verbatim}
\usepackage{makecell}
\usepackage{tabularx}
\usepackage{algorithm}
\usepackage{algpseudocode}
\usepackage{hhline}
\usepackage{booktabs}
\usepackage{eso-pic}

\usepackage{censor}

\title{\LARGE \bf
Control Architecture for Safe Grasping of Fragile Objects \\ 
Using a Coarse Position-Controlled Gripper}

\author{Marko Pavlic, Moritz Geier, Timo Markert, and Darius Burschka%
\thanks{The authors acknowledge the financial support by the Munich Institute of Robotics and Machine Intelligence (MIRMI).
}
\thanks{Marko Pavlic, Moritz Geier and Darius Burschka are with the Machine Vision and Perception Group, Chair of Robotics, Artificial Intelligence and Real-time Systems, Technical University of Munich, Munich, Germany.}%
\thanks{Timo Markert is with Resense GmbH, Klingenberg, Germany and with the Semantic Information Systems Group, Osnabrück University, Osnabrück, Germany.}%
\thanks{Corresponding author: Marko Pavlic, {\tt\footnotesize marko.pavlic@tum.de}.}%
}

\begin{document}

\AddToShipoutPictureBG*{%
  \AtPageUpperLeft{%
    \raisebox{-1.0cm}{%
      \makebox[\paperwidth][c]{%
        \parbox{\textwidth}{\centering\footnotesize
          \copyright\,2026 IEEE. Personal use of this material is permitted. Permission from IEEE must be obtained for all other uses, in any current or future media, including reprinting/republishing this material for advertising or promotional purposes, creating new collective works, for resale or redistribution to servers or lists, or reuse of any copyrighted component of this work in other works.}%
      }%
    }%
  }%
}

\maketitle
\thispagestyle{empty}
\pagestyle{empty}

\begin{abstract}
Robots are increasingly used in unstructured environments. The need for them to safely grasp unknown objects without damaging them becomes crucial. Humans achieve this by sensing and quickly responding by adjusting their grasping force. Similarly, effective grasp acquisition in robots requires compliant interaction strategies that can adapt to uncertain object properties and adjust to any instabilities during manipulation.

We present a geometry-aware force/torque-based contact estimation method for a coarse position-controlled gripper, combined with an adaptive admittance controller for safe grasp acquisition. The desired contact forces are estimated online to keep stable contact with objects of unknown properties. This enables compliant and stable grasps while avoiding excessive forces. Experiments with objects of different sizes, shapes, stiffnesses, and weights show that the proposed algorithm not only prevents slippage but also applies minimal force to safely grasp an object without causing excessive deformation.
\end{abstract}

\section{Introduction}
\label{sec:intro}

Robotic grasping in unstructured environments remains a challenging problem due to unknown or uncertain object properties, such as geometry, material properties, and contact conditions. But also because gripper hardware is often not designed for these scenarios. Humans can adapt quickly to instability through tactile sensing and adjust their grasp to avoid dropping the object while keeping it intact. Tactile perception provides critical information about the object’s physical properties and the contact state between the object and the gripper. However, grasping force control is currently mostly designed and adapted to specific, well-defined use cases. Most existing approaches assume prior knowledge of the object’s properties, enabling the controller to be designed for predictable interaction conditions. However, these methods often struggle to generalize to unstructured environments, where robots must handle a wide variety of objects with unknown and potentially varying physical characteristics. In such scenarios, there is a need for more adaptive and generalizable force control strategies.

\begin{figure}
    \centerline{\includegraphics[width=\columnwidth]{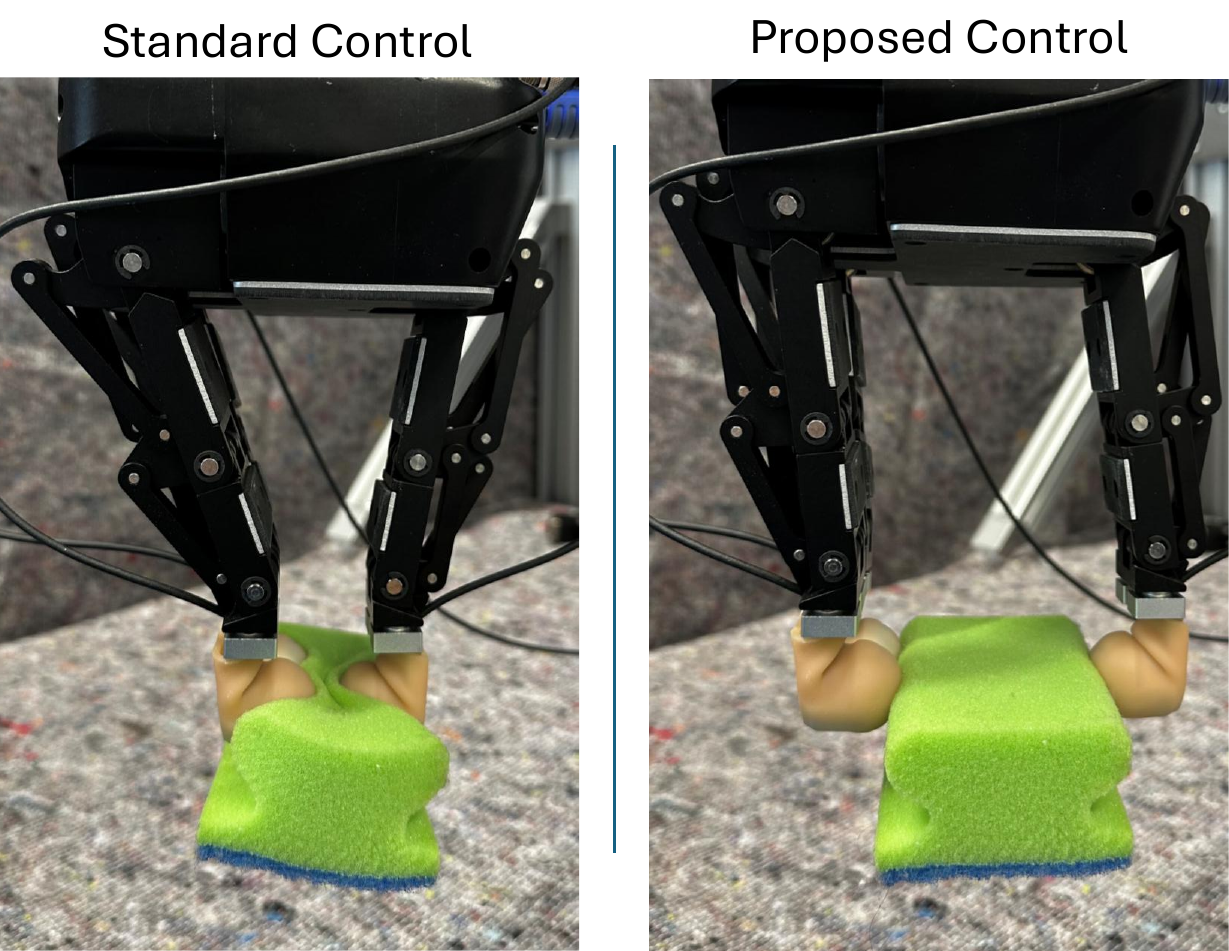}}
    \caption{\textbf{Safe grasping of deformable objects.} In industrial and research settings, widely used position-controlled grippers perform badly on soft and deformable objects. Our proposed control scheme uses force/torque (F/T) sensors in the fingertips to calculate the minimum force required for a stable grasp of any unknown object.}
    \label{fig:teaser}
\end{figure}

While position-controlled grippers are widely used in industrial and research settings, pure position tracking during contact can lead to unstable grasps or excessive forces that damage objects (see Figure \ref{fig:teaser}). Although these grippers can limit the maximum grasp force through current control, high-fidelity force control is not possible. Therefore, compliant control strategies on top of the internal position controller are essential for safe interaction with unknown objects.

A common approach to compliant interaction is admittance control, in which measured contact forces are mapped to motion commands via a virtual mass–spring–damper system. However, selecting appropriate admittance parameters typically requires prior knowledge of object properties or extensive tuning, which limits applicability in unknown environments.

We address safe grasp acquisition using fingertip force/torque (F/T) sensing and online estimation of the desired minimal contact forces, combined with an admittance controller that maps force commands to motion commands.

During grasp acquisition, the robot must establish contact safely while adapting its compliance to unknown object characteristics. This motivates the geometry-aware estimation of contact states and the online adaptation of the desired grasping. The latter is achieved by continuously monitoring the tangential force components, which serve as primary indicators of incipient slip. Once a grasp has been established, the challenge shifts to maintaining it, as contact conditions may change due to object motion, disturbances, or internal grasp adjustments.

Our contributions are summarized as follows:
\begin{enumerate}
    \item We propose a geometry-aware contact estimation method using fingertip F/T sensing.
    \item We estimate the minimal slip-preventing grasp force from tangential contact forces instead of relying on prior knowledge of object properties.
    \item We develop an adaptive control strategy for force-regulated grasping with a position-controlled gripper.
\end{enumerate}

The rest of the paper is organized as follows: \sref{sec:relatedWork} discusses related work before the system design, and the proposed approach is explained in detail in \sref{sec:F/Tcontrol}, followed by the experiments and results in \sref{sec:results}.
Finally, \sref{sec:conclusion} concludes with a summary and outlook on future directions.

\section{Related Work}
\label{sec:relatedWork}

Compliance control has been extensively studied as a key mechanism for enabling safe and robust physical interaction in robotic manipulation. Sadun \textit{et al.}~\cite{sadun_overview_2016} provided a comprehensive overview of compliance control strategies, distinguishing between passive and active approaches and highlighting their respective advantages in handling contact uncertainty. It was mentioned that contact-state modeling is key to a controller's capability. Further explored by Sadun \textit{et al.} in \cite{sadun_adaptive_2015}, highlighted the importance of friction modeling in compliant grasping. These studies underscore that accurate interaction modeling significantly influences force control performance.

Early work on compliant control focused on using low-cost FSR sensors and simple force control strategies. Sadun \textit{et al.}~\cite{sadun_force_2016} implemented fingertip force sensing combined with a proportional force control law layered on a PID position controller. Although effective in principle, the study reports performance degradation with low-cost sensors, leading to noisy, corrugated force signals.

Huynh and Kuo~\cite{huynh_optimal_2020} proposed an admittance-based impedance controller for robotic grasping, demonstrating improved interaction stability via fuzzy-logic optimization. Their approach shares the objective of controlling contact behavior using a position-controlled gripper. However, controller performance remains dependent on rule-based tuning and task-specific parameter selection, such as knowledge of the object's weight and friction coefficient.

Operational-space compliance control has been investigated by Jalani \textit{et al.}~\cite{jalani_active_2013}, who introduced an Integral Sliding Mode Control framework for active hand compliance. While the controller parameters provide strong robustness, they also require tuning for specific object classes.

So, in summary, most conventional active compliance frameworks rely on manually specified target forces, knowledge of object properties, or tuning dependency for specific object classes, which limits their applicability to known objects. This dependency motivates different approaches, which seek to infer appropriate control behavior directly from contact measurements, eliminating the need for object-specific parameterization.

\begin{table*}[t]
\centering
\caption{Comparison with related force-regulated grasping approaches.}
\label{tab:comparison}
\begin{tabularx}{\textwidth}{@{}l X X X X@{}}
\toprule
\textbf{Method} & \textbf{Platform} & \textbf{Sensing} & \textbf{Object assumptions} & \textbf{Approach} \\
\midrule
O'Toole et al.\ \cite{OToole.2010} &
Simulation + custom test rig &
Carriage position; no force sensor &
None assumed &
Force inferred via observer; slow to converge, sensitive to model--plant mismatch \\
Sadun et al.\ \cite{sadun_adaptive_2015} &
Simulation only (single joint) &
None (model-based) &
Explicit friction / mass--spring--damper model &
Model reference adaptive compliance control \\
Ding et al.\ \cite{ding_adaptive_2019} &
Custom 2-finger prototype &
Single FSR (one finger) + laser optical slip sensor (other finger) &
None assumed &
Adaptive regrasping triggered after slip is detected \\
Al-Mohammed et al.\ \cite{al-mohammed_switched_2023} &
Custom 2-finger prototype &
Single FSR (one finger) + two laser slip sensors (other finger) &
Medium--high stiffness (rigidity assumption) &
Switched adaptive control for translational + rotational slip \\
\textbf{Ours} &
Commercial coarse position-controlled gripper &
F/T sensor at each fingertip &
None assumed &
Desired force from tangential-force feedback\\
\bottomrule
\end{tabularx}
\end{table*}

Several works regulate grasp force toward the minimum required to prevent slippage. Ding \textit{et al.}~\cite{ding_adaptive_2019} and Al-Mohammed \textit{et al.}~\cite{al-mohammed_switched_2023} proposed adaptive control algorithms for grasping unknown objects on a custom two-finger sensorized prototype, using a single force-sensing resistor on one finger and laser-based optical slip sensor(s) on the other to adjust grasp force in response to detected slip. Ding \textit{et al.}\ demonstrated their approach on genuinely delicate objects, including empty Styrofoam and paper cups, showing reduced deformation compared to open-loop and hardware-limited grasping. Al-Mohammed \textit{et al.}\ extended this design to compensate for rotational as well as translational slip, but explicitly restrict their method to objects of medium-to-high stiffness, excluding highly deformable objects by design. O'Toole \textit{et al.}~\cite{OToole.2010} addressed a related problem using a fuzzy sliding-mode controller combined with a disturbance observer, validated on both simulation and a physical test rig. The authors note that their observer requires low-pass filtering and is sensitive to model--plant mismatch, so its estimate is slow to converge, and optimal performance is not guaranteed. Sadun \textit{et al.}~\cite{sadun_adaptive_2015} instead take a model-based route with an explicit friction and mass--spring--damper model, validated only in simulation on a single joint.

These approaches differ from the proposed one in how contact force is obtained and on what hardware. Ding \textit{et al.}\ and Al-Mohammed \textit{et al.}\ rely on a single-point FSR on only one finger together with dedicated slip sensors, rather than distributed force/torque sensing at every contact point; Al-Mohammed \textit{et al.}\ further assume sufficient object rigidity, an assumption our method does not require. Furthermore, their regrasping controller is activated only upon detection of a physical slip event, whereas ours continuously updates the desired grasp force based on tangential force feedback, allowing it to act before slippage occurs rather than reactively afterward. 

The proposed method regulates the grasp using directly measured, high-precision force/torque at each fingertip. It integrates contact-state estimation, geometry-aware force distribution, and adaptive control. The desired grasping forces are adjusted online solely based on measured interaction signals, enabling stable grasp acquisition of objects with unknown properties without manual parameter tuning, while running on a standard, coarse-position-controlled industrial gripper. A direct experimental comparison with the related works presented is not feasible due to differing hardware and sensing requirements; instead, Table~\ref{tab:comparison} summarizes the structural differences discussed above.

\section{Force/Torque-Based Contact Estimation and Control}
\label{sec:F/Tcontrol}

The goal of this research is to design an algorithm for safe grasp acquisition of unknown objects, despite the hardware limitations of widely used position-controlled grippers in industry and research. We want to apply the minimum force needed to prevent the object from slipping while maintaining the deformation constraint. For this, we use fingertip-mounted F/T sensors. The framework pipeline combines signal filtering, contact geometry estimation, geometry-aware force distribution, and adaptive admittance control to regulate contact forces between the gripper and the object while avoiding damage to the object.

\subsection{Control Framework}
\label{sec:control_framework}

\begin{figure}
     \centerline{\includegraphics[width=\columnwidth]{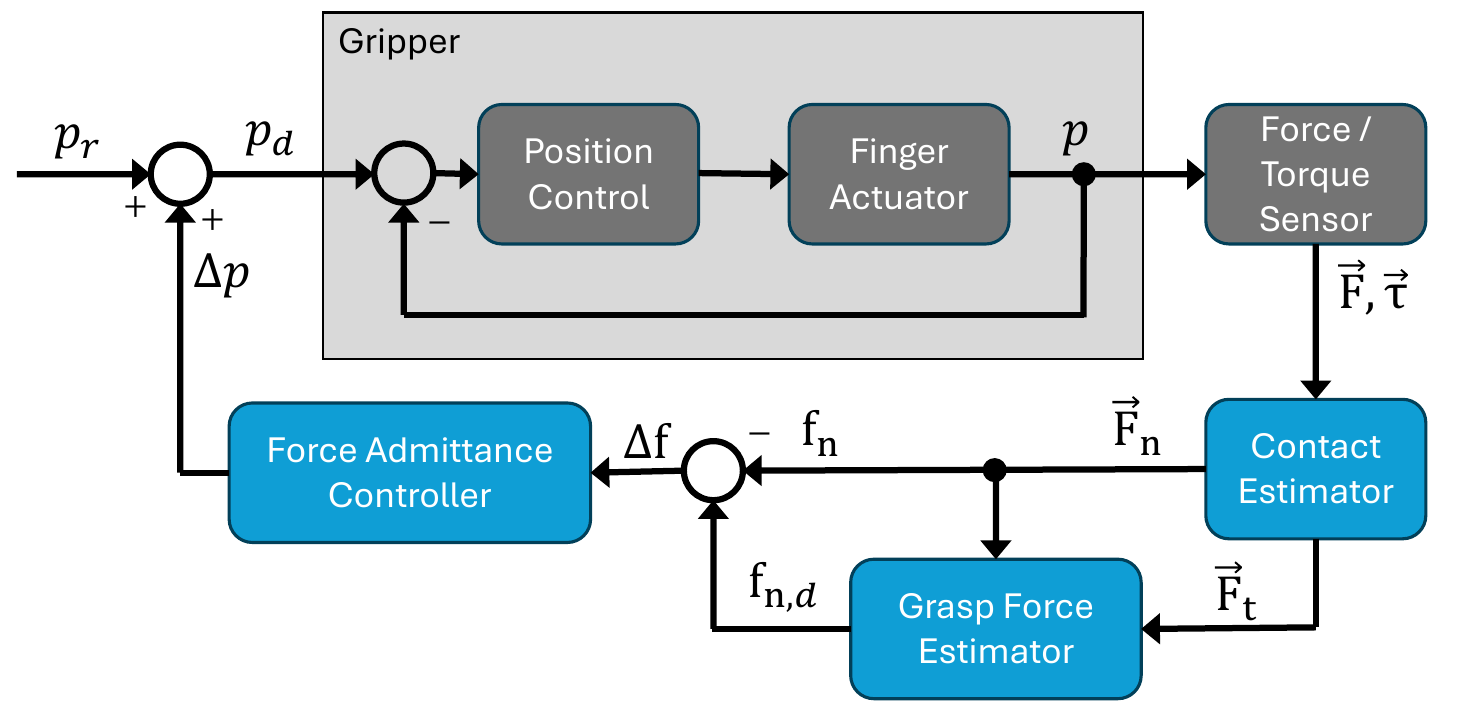}}
    \caption{\textbf{Control framework.} Proposed control architecture for adaptive grasping of a variety of unknown objects. Based on fingertip F/T measurements, the proposed system adapts the gripper's position controller commands. Highlighted in blue are the contributed modules.}
    \label{fig:controlArchitectur}
\end{figure}

Figure \ref{fig:controlArchitectur} shows the complete control framework for the proposed adaptive grasp strategy designed in this work. For objects with unknown shape and size, the position reference $p_r$ is initially set to a fully closed position and updated upon contact with the object. The F/T measurements from the fingertips are fed into the \textit{Contact Estimator} to calculate the contact states $f_n$ and $f_t$, which are further processed in the \textit{Grasp Force Estimator} to derive the desired contact force $f_{n,d}$. Based on the force deviation, the admittance controller calculates the position compensation $\Delta p$ for the gripper, which is added to the reference position $p_r$ to adjust the gripper position so that the desired force is exerted on the object.

\subsection{Signal Filtering}

The measured wrench at the sensor frame is given by
\begin{equation}
\mathbf{w} =
\begin{bmatrix}
\mathbf{F} \\
\mathbf{\tau}
\end{bmatrix},
\label{eq:FT}
\end{equation}
where $\mathbf{F} \in \mathbb{R}^3$ is the contact force and $\mathbf{\tau} \in \mathbb{R}^3$ is the torque about the sensor origin. These measurements are subject to high-frequency noise, and to ensure reliable contact detection and force estimation, all raw signals are filtered using a low-pass filter prior to further processing. Each component of the measured force and torque vectors is filtered independently using a Butterworth \cite{Butterworth1930} low-pass filter. The Butterworth filter was chosen for its maximally flat passband magnitude response and absence of ripple, both of which are desirable for contact force estimation.

The cutoff frequency $f_c$ is selected based on the expected bandwidth of contact forces during grasping \cite{johansson_coding_2009, Romano2011}. Since grasp forces evolve relatively slowly compared to the sensor sampling rate $f_s = 1000~\mathrm{Hz}$, a cutoff frequency of $f_c = 50~\mathrm{Hz}$ was found to effectively suppress noise while preserving contact dynamics. For simplicity, we will not distinguish between raw and filtered signals in the remainder of this document and will treat all measurements as filtered with the same low-pass filter.

\subsection{Geometry Constrained Contact Point Estimation}
\label{sec:contact_estimation}

\begin{figure}[tbp]
    \centerline{\includegraphics[width=\columnwidth]{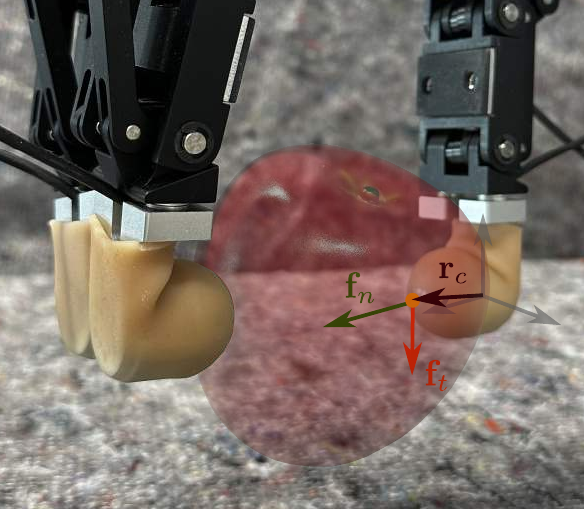}}
    \caption{\textbf{Contact state estimation.} Estimated contact states (contact point, normal and tangential force components) during interaction between fingertip and object. Instead of relying on object properties, the grasping force is adapted directly from measured contact interactions.}
    \label{fig:contact-state}
\end{figure}

In our proposed approach, neither the object mass nor the friction coefficient is assumed to be known. Instead of relying on an explicit contact model, the grasping force is adapted directly from measured contact interactions.

After contact with the object is established, the wrench measurements are used to estimate the contact point between the fingertips and the object.

For contact state estimation, not only are the six-axis F/T measurements used, but also the fingertip geometry model. A hemispherical-shaped fingertip with radius $R$ is attached to the F/T sensor. Contact is detected when the measured force magnitude exceeds a threshold of $f_{\mathrm{th}}$. At initial contact, torque measurements are typically noisy and unreliable due to low force magnitudes. In that situation, the contact point $\mathbf{r}_c$ is approximated using a force-only estimate of the contact normal $\mathbf{n}$ and the projection of it onto the hemispherical fingertip surface as
\begin{equation}
\mathbf{r}_c = - R \frac{\mathbf{F}}{\|\mathbf{F}\|} = R\,\mathbf{n}.
\label{eq:force_only_normal}
\end{equation}

Once the contact force exceeds a second predefined switch threshold $f_{\mathrm{sw}}$, torque measurements become sufficiently informative to refine the contact localization. In this case, a particular solution can be obtained by
\begin{equation}
\mathbf{r}_0 = \frac{\mathbf{F} \times \mathbf{\tau}}{\|\mathbf{F}\|^2},
\label{eq:wrench_contact}
\end{equation}

However, \(\mathbf{r}_0\) represents only one point along the line of action of the force. The true contact location lies along

\begin{equation}
\mathbf{r}(\lambda) = \mathbf{r}_0 + \lambda \mathbf{F},
\label{eq:force_line}
\end{equation}

with scalar parameter \(\lambda \in \mathbb{R}\). Since contact occurs somewhere on the hemispherical fingertip of radius $R$, the contact point must satisfy

\begin{equation}
\|\mathbf{r}(\lambda)\|^2 = R^2.
\label{eq:sphere_constraint}
\end{equation}

Substituting \eqref{eq:force_line} into \eqref{eq:sphere_constraint} yields in the quadratic equation

\begin{equation}
\|\mathbf{r}_0\|^2 + 2\lambda (\mathbf{r}_0 \cdot \mathbf{F}) + \lambda^2 \|\mathbf{F}\|^2 = R^2,
\label{eq:r_quadratic}
\end{equation}

Solving \eqref{eq:r_quadratic} produces two candidate solutions ($\lambda_{1}, \lambda_{2}$) corresponding to the intersections of the force line with the spherical surface. The physically valid contact point is selected based on

\begin{equation}
\mathbf{r}_c =
\mathbf{r}(\lambda_i)
\quad
\text{s.t.}
\quad
\mathbf{r}(\lambda_i)^\top \mathbf{F} < 0
\qquad i \in \{1,2\}
\label{eq:contact_selection}
\end{equation}

which enforces that the measured force acts inward on the hemispherical surface. The contact normal is defined as the outward surface normal at the estimated contact point,
\begin{equation}
\mathbf{n} = \frac{\mathbf{r}_c}{\|\mathbf{r}_c\|}.
\end{equation}

Finally, the measured force is decomposed into a normal and tangential component as
\begin{equation}
f_n = \mathbf{F}^\top \mathbf{n}, \qquad
\mathbf{f}_t = \mathbf{F} - f_n \mathbf{n}.
\label{eq:force_decomposition}
\end{equation}

The contact state estimation is visualized in \fref{fig:contact-state}.

We adopt a hard-finger point-contact model \cite{Salisbury1983KinematicAF}. Physically, the rubber cover forms a finite contact patch (a soft-finger contact); if the grasped object rotates about the contact normal, the resulting torsional moment is measured by the F/T sensor but is not represented in the hard-finger model, and would therefore be partially absorbed into the estimated contact point and force components. The cover thickness ($\approx 1~mm$) is small relative to the fingertip radius ($R=12~mm$), so the resulting contact patch — and hence the torsional moment it can transmit — is negligible compared to the resolved normal and tangential forces.

\subsection{Slip--Driven Normal Force Adaptation}

After contact with the object was established and a first estimation of normal and tangential force is available, the lifting phase of the robot is initiated. In our proposed approach, neither the object mass nor the friction coefficient is assumed to be known. Instead of relying on an explicit contact model, the grasping force is adapted directly from measured contact interactions.

Slip tendency is inferred from the temporal evolution of the
tangential force magnitude. The tangential force rate is estimated using a discrete derivative:

\begin{equation}
\dot{f}_t =
\frac{\|\mathbf{f}_t(k)\| - \|\mathbf{f}_t(k-1)\|}
{\Delta t}
\label{eq:ft_dot}
\end{equation}
where $\Delta t$ is the sampling time.
An increase in $\|\mathbf{f}_t\|$ indicates growing shear interaction between fingertip and object, which typically precedes slipping. Based on that, a slowly varying baseline normal force $f_{n,0}$ is introduced to capture the minimum force required so that no slippage occurs. The baseline is updated according to

\begin{equation}
f_{n,0}(k) =
f_{n,0}(k-1) + k_i \, \dot{f}_t \, \Delta t
\label{eq:baseline_update}
\end{equation}

where $k_i > 0$ is an adaptation gain. This mechanism increases the baseline force when slip tendency is detected, while remaining constant under steady-contact conditions. As a fixed controller gain, $k_i$ sets the rate of this adaptation and does not depend on the object's mechanical properties. To enforce minimal contact with the object, the baseline force is bounded $f_{n,0}(k) \leftarrow \max\!\left(f_{n,\min}, \, f_{n,0}(k)\right)$. The minimum force $f_{n,min}$ required to maintain contact is determined by the sensor noise level.

The desired normal force, which is supplied to the admittance controller, is then defined as

\begin{equation}
f_{n,\mathrm{d}} =
f_{n,0} + k_s \|\mathbf{f}_t\|
\label{eq:fn_des}
\end{equation}

where $k_s > 0$ is a proportional scaling factor, which introduces a direct coupling between tangential loading and grasping force. This term captures slower adaptation to the additional tangential load by the object weight during lifting, before slippage occurs. Like $k_i$, it is a fixed gain that shapes the controller's response rather than encoding any object-specific parameter. To not overload the F/T sensor and the gripper actuators, the desired normal force is bounded $f_{n,\mathrm{d}} \leftarrow \min\!\left(f_{n,\max}, \, f_{n,\mathrm{d}}\right)$.

The proposed force estimator, therefore, implements an artificial load-force coupling mechanism that stabilizes contact without requiring estimation of object mass or friction parameters.

\subsection{Adaptive Admittance Control}
\label{sec:admittance_stiffness}

In this work, finger closure is regulated using an admittance controller. The generalized position variable $\Delta p$ corresponds to the incremental finger displacement command, while interaction forces are measured at the fingertips. The admittance dynamics are given by
\begin{equation}
M \ddot{\Delta p} + D \dot{\Delta p} + K \Delta p = f_{n,\mathrm{d}} - f_n,
\label{eq:admittance_full}
\end{equation}
where $M$, $D$, and $K$ represent inertia, damping, and stiffness parameters, respectively. $f_n$ is the measured normal contact force from \eqref{eq:force_decomposition} and $f_{n,\mathrm{d}}$ is the desired normal force obtained by \eqref{eq:fn_des}. The admittance output $\Delta p$ is an incremental command applied to the current finger position $p_r$, giving the commanded position $p_d = p_r + \Delta p$. The gripper accepts positions on a discrete integer scale, so $p_d \in [0,255]$ and correspondingly $\Delta p \in [-255, 255]$, where positive values close and negative values open the gripper. The admittance dynamics in \eqref{eq:admittance_full} are implemented as an outer control loop that generates a position reference for the gripper's internal position controller.

\subsection{Stability Analysis}
As is standard in force-regulating admittance control, we model the
local contact as an elastic spring and neglect the environment mass and damping
\cite{Liu_Li_2023,al-mohammed_switched_2023}. The normal force is proportional to the normal indentation $\delta\ge 0$ of the spherical fingertip into the object,
\begin{equation}
f_n = K_e\,\delta ,
\label{eq:contact_spring}
\end{equation}
with $K_e>0$ the (unknown, bounded) contact stiffness.
The admittance \eqref{eq:admittance_full} commands motion $\Delta p$ along the
fixed gripper closing direction $\mathbf e_z$, whereas the estimated contact normal $\mathbf n$ generally deviates from it by an angle $\theta$. The
indentation produced by a displacement $\Delta p$ along $\mathbf e_z$ is its
projection onto the normal,
\begin{equation}
\delta = (\mathbf n^{\top}\mathbf e_z)\,\Delta p = n_z\,\Delta p,
\qquad n_z := \cos\theta \in (0,1],
\label{eq:projection}
\end{equation}
so that
\begin{equation}
f_n = K_e\, n_z\,\Delta p .
\label{eq:contact_nz}
\end{equation}

Following the timescale-separation methodology of \cite{al-mohammed_switched_2023}, the saturated force-reference law \eqref{eq:fn_des} is slow relative to the admittance \eqref{eq:admittance_full} and acts as a constant compliance set-point $f_{n,\mathrm{d}}$ for the inner loop; the objective is convergence of $\Delta p$ to its rendered equilibrium, not force tracking, since $K>0$ leaves an intended steady-state offset.

Substituting \eqref{eq:contact_nz} into \eqref{eq:admittance_full} yields the inner closed loop
\begin{equation}
  M\,\ddot{\Delta p} + D\,\dot{\Delta p} + (K+K_e\, n_z)\,\Delta p = f_{n,\mathrm{d}},
  \label{eq:coupled}
\end{equation}
with $M,D>0$, $K\ge0$, and unique equilibrium
$\Delta p^\star = f_{n,\mathrm{d}}/(K+K_e\, n_z)$. With the error
$e \triangleq \Delta p - \Delta p^\star$ and the energy function
\begin{equation}
  V = \tfrac12 M\,\dot e^{\,2} + \tfrac12 (K+K_e\, n_z)\,e^{2} \;\ge\; 0,
  \label{eq:lyap}
\end{equation}
differentiation along \eqref{eq:coupled} gives
\begin{equation}
  \dot V = -\,D\,\dot e^{\,2} \;\le\; 0.
  \label{eq:vdot}
\end{equation}
As $V$ is positive definite and radially unbounded with
$\dot V\le 0$, the standard Barbalat/LaSalle argument \cite{al-mohammed_switched_2023}
yields $(e,\dot e)\to(0,0)$; hence $\Delta p\to\Delta p^\star$ and the contact
force settles to the rendered value
$f_n^\star=\tfrac{K_e\, n_z}{K+K_e\, n_z}\,f_{n,\mathrm{d}}\le f_{n,\mathrm{d}}$, whose
intrinsic offset bounds the applied force for any object stiffness and thereby
protects fragile grasps.

The result holds for every $K_e\, n_z>0$. This positivity is guaranteed, since the contact is unilateral and is established by the closing motion itself, so a contact can only form and be loaded on the leading hemisphere of the spherical fingertip, for which $n_z=\cos\theta>0$. A larger $K_e$ only raises the admittance bandwidth
$\omega_{\mathrm{adm}}=\sqrt{(K+K_e\, n_z)/M}$, so the stiffest object $K_{e,\max}$ is the binding case for the timescale separation; $M$ is sized so that
$\omega_{\mathrm{adm}}(K_{e,\max})\ll\omega_{\mathrm{pos}}$, the internal
position-controller bandwidth.

\section{Experiments and Results}
\label{sec:results}

\begin{figure*}[!t]
\centerline{\includegraphics[width=\textwidth]{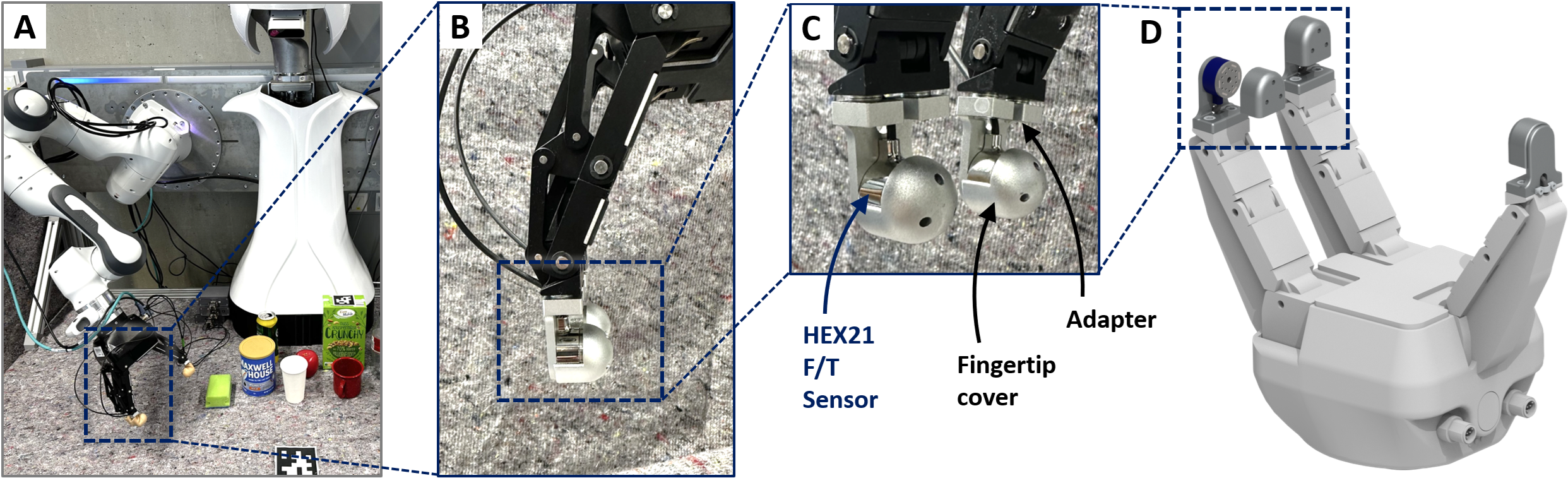}}
\caption{\textbf{Robot setup.} The grasping trials are conducted on a bimanual manipulation system comprising two Franka robotic arms. One arm carries a Robotiq 3-Finger Gripper instrumented with miniature 6-axis F/T sensors integrated directly into the fingertips for contact force measurement. 
\textbf{A} Overview of the robot setup with a representative subset of the grasped objects. 
\textbf{B} Detailed view of the gripper end-effector highlighting the integrated miniature sensors and cable routing. 
\textbf{C} Close-up of the instrumented robot fingertips featuring integrated Resense HEX21 6-axis F/T sensors, custom-designed adapter parts, and rigid hemispherical fingertips. 
\textbf{D} CAD rendering of the Robotiq 3-Finger Adaptive Gripper equipped with sensorized fingertips.}
\label{fig:robot-setup}
\end{figure*}

All experiments are conducted on a bimanual manipulation platform shown in \fref{fig:robot-setup}. One of the Franka arms carries a position-controlled Robotiq 3-Finger Gripper, which can set %
rough force limits from ($15~N - 60~N$). We equip the grippers' fingertips with Resense HEX21 6-axis F/T sensors in order to measure interaction forces directly at the point of contact. The fingertip integration is inspired by \cite{Pavlic.2023}, where a similar hardware setup is used to estimate physical properties of unknown objects. The F/T sensors provide three-dimensional force and torque measurements in the local sensor frame. To establish a human-finger-like contact with the object, an aluminium-sphere fingertip was designed and coated with a thin layer of rubber to provide additional friction. As test objects, we used common, delicate, and deformable household items of varying weights, and we also tested the framework with stiff objects to demonstrate the proposed approach's versatility.

The control parameters for the proposed framework are listed in \tref{tab:controller_parameters} and were tuned once for the proposed system and held constant across all experiments and object types.

\begin{table}
\centering
\caption{Controller Parameters}
\label{tab:controller_parameters}
\begin{tabular}{llll}
\hline
\textbf{Symbol} & \textbf{Description} & \textbf{Value}\\
\hline

$k_i$ & Slip adaptation gain & 3.0 \\

$k_s$ & Tangential force coupling gain & 3.0 \\

$f_{\min}$ & Minimum normal force & 0.6 \\

$f_{\max}$ & Maximum normal force & 40.0 \\

$M$ & Virtual mass (admittance) & 0.7\\

$K$ & Virtual stiffness (admittance) & 30.0\\

$D$ & Virtual damping (admittance) & 100.0 \\

$\Delta t$ & Outer (admittance) control sampling time & 0.001\\

$\Delta t_{int}$ & Internal (position) control sampling time & 0.01\\

\hline
\end{tabular}
\end{table}

\subsection{Safe Grasp Acquisition with Unknown Objects}

\begin{figure}
    \centerline{\includegraphics[width=0.9\columnwidth]{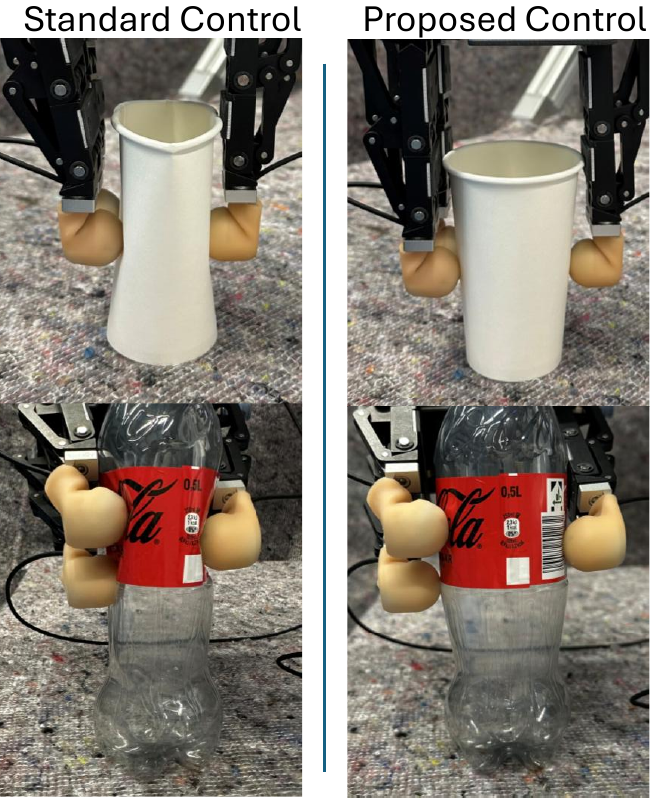}}
    \caption{\textbf{Comparison between standard and adaptive control.} Different delicate objects being grasped with the gripper's internal controller (left) compared to the proposed grasping algorithm (right).}
    \label{fig:posControlComparison}
\end{figure}

As discussed in Section~\ref{sec:relatedWork}, a direct experimental comparison with prior adaptive grasping methods is not meaningful here, as each depends on different actuation and sensing hardware, which was compared and summarized in Table~\ref{tab:comparison}.
In a first experiment, we compared the gripper's internal grasping control with the proposed framework on the same actuation platform, so that the comparison reflects the combined effect of the proposed sensing and control framework against the gripper's default capability. The gripper provides a coarse position controller with contact detection based on force estimates derived from motor current measurements. This built-in contact detection has a minimum configurable threshold of $15~N$; we therefore set it to this lowest possible value. The gripper was closed until object contact was detected. We compared grasping performance on deformable objects between the gripper's own contact detection (referred to as \textit{Standard Control} in this section) and our proposed controller (\textit{Proposed Control}). \fref{fig:posControlComparison} shows the visual result for two objects and clearly reveals the deformation of the test objects during the grasp with the \textit{Standard Control}, whereas the \textit{Proposed Control} stops at the object's shape without noticeable deformation.

\begin{figure}
    \centerline{\includegraphics[width=\columnwidth]{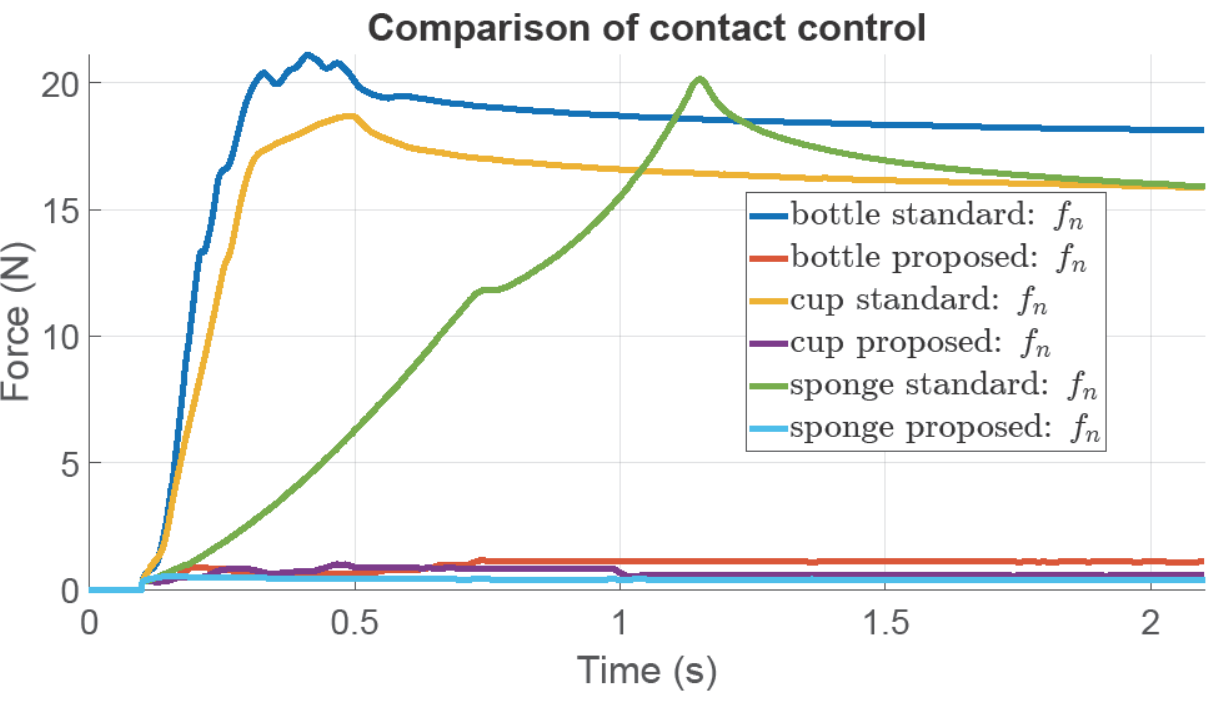}}
    \caption{\textbf{Contact force comparison.} Three different deformable objects (plastic bottle, paper cup, and sponge) are grasped with the gripper's internal controller (standard) and the proposed adaptive grasping algorithm. The graph shows the applied grasping forces, which are an order of magnitude smaller for the proposed control.}
    \label{fig:posControlComparisonPlot}
\end{figure}

The measured normal forces $f_n$ from a single finger are depicted in \fref{fig:posControlComparisonPlot} and confirm the visual observations. 
The forces under \textit{Standard Control} settle slightly above the set value of $15~N$, consistent with the gripper estimating contact force from motor current, which is less accurate than a calibrated F/T sensor. Crucially, even at its lowest achievable threshold, this force level exceeds what delicate deformable objects, such as paper cups, plastic bottles, or sponges, can withstand, resulting in visible damage. In contrast, the \textit{Proposed Control} regulates the grasp well below this floor on the same actuation, using feedback from the fingertip F/T sensors.

These results demonstrate that the limitation is not merely a matter of tuning but is inherent to the standard current-based contact detection for this class of coarse position-controlled grippers: without additional sensors or prior knowledge of the object's mechanical properties, such grippers cannot grasp delicate, deformable objects without damaging them. The proposed framework overcomes this limitation on the same actuation platform by combining fingertip F/T sensing with an admittance controller.

\subsection{Force Adaptation During Lift}

In the second experiment, the proposed adaptive force control during object lifting is investigated. When initiating a grasp of an object, we assume the object is supported, and close the gripper until contact is detected, at which point the adaptive force control is automatically activated and exerts a minimum force of $f_{min} = 0.6~N$. During lifting, the adaptive controller framework continuously observes F/T measurements and adapts to any slippage or a tangential force increase due to weight, successfully grasping the object with minimal required force.

\begin{figure}
    \centerline{\includegraphics[width=\columnwidth]{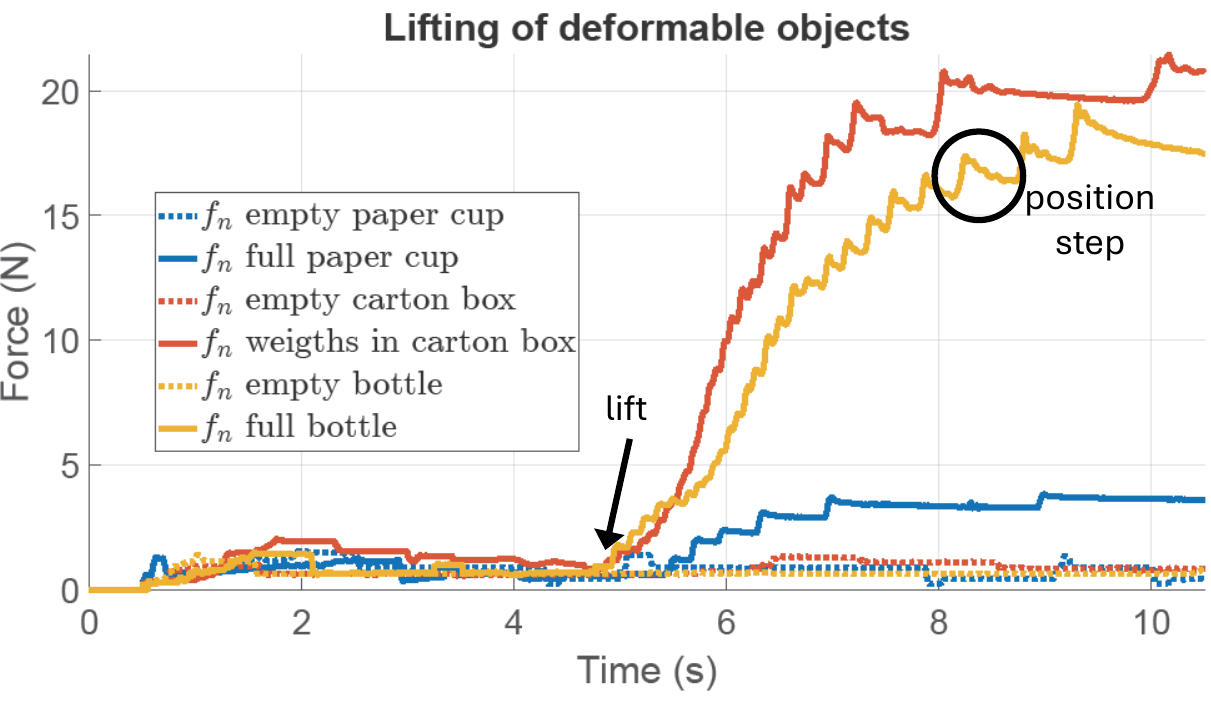}}
    \caption{\textbf{Lifting deformable objects.} Measured normal forces during the lifting of different deformable objects with varying weights.}
    \label{fig:liftPlot}
\end{figure}

When lifting deformable objects, it is necessary to keep the applied force to a minimum to avoid damaging the object. In \fref{fig:liftPlot}, the measured normal forces during the lifting of different objects are compared. After establishing contact at $\approx 0.5~s$, the lift event starts at $\approx 5.0~s$. Since all the deformable objects under investigation are very light when empty, there is little deviation from the initial contact force when lifting them, which is shown by the dashed lines. The initial contact force is already sufficient to lift it successfully. The gripper's response becomes more pronounced when additional weight is applied, requiring it to exert greater force on the objects. This was achieved by filling the paper cup and plastic bottle with water and adding weights in the carton box. The measured normal forces for these experiments are shown in solid lines and clearly show the force adjustments due to the increased weight. The paper cup and plastic bottle were only minimally deformed during this experiment, which we considered acceptable given the relatively small contact areas between the fingers and the objects. It has to be noted that the plastic bottle was kept open during the experiment to make deformation easier and the task more challenging.

\begin{figure}
    \centerline{\includegraphics[width=\columnwidth]{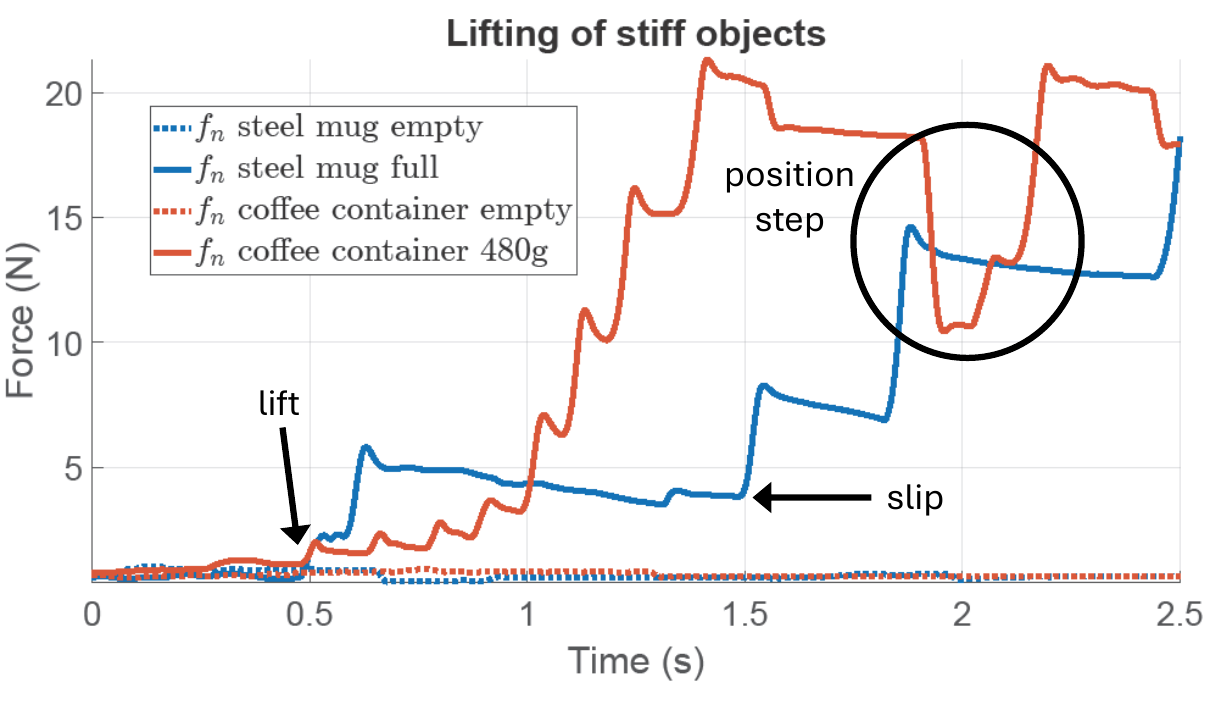}}
    \caption{\textbf{Lifting stiff objects.} Measured normal forces during the lifting of stiff objects with varying weights. The same control parameters that perform well for deformable objects also work well for stiff objects.}
    \label{fig:liftStiffPlot}
\end{figure}

To demonstrate that the same controller parameters also work for other objects, the same experiments were also performed for rigid objects. The measured normal forces are depicted in \fref{fig:liftStiffPlot}. The dashed lines again indicate that the object is empty during the lift, and the solid lines show the experiment with an added weight ($480~g$) to the coffee container and the steel mug filled with water. Both objects were lifted successfully, with the steel mug showing a slight slip after some time, which was corrected by increasing the force without dropping the cup.

In both \fref{fig:liftPlot} and \fref{fig:liftStiffPlot}, the measured normal force shows a stepwise profile. This results from the gripper's coarse position resolution described in Section~\ref{sec:admittance_stiffness}, so each position increment produces a discrete jump in contact force rather than a continuous evolution. The force steps become more pronounced for stiffer objects, since a fixed position increment translates into a larger force change at higher contact stiffness. The brief force transient before each step settles reflects the viscoelastic response of the rubber fingertip cover, which introduces compliance and damping at the contact.

Nevertheless, the proposed control scheme successfully grasps objects of varying physical properties without requiring prior knowledge of their characteristics or controller retuning. Furthermore, the approach prevents damage to deformable objects during grasping. The applied force need not match $f_{n,\mathrm{d}}$ exactly: since no object parameters are assumed, it suffices that the force falls within the object's safe range, as confirmed by the absence of visible deformation across all tested objects.

\subsection{Force Adaptation During Disturbance}

\begin{figure}
    \centerline{\includegraphics[width=\columnwidth]{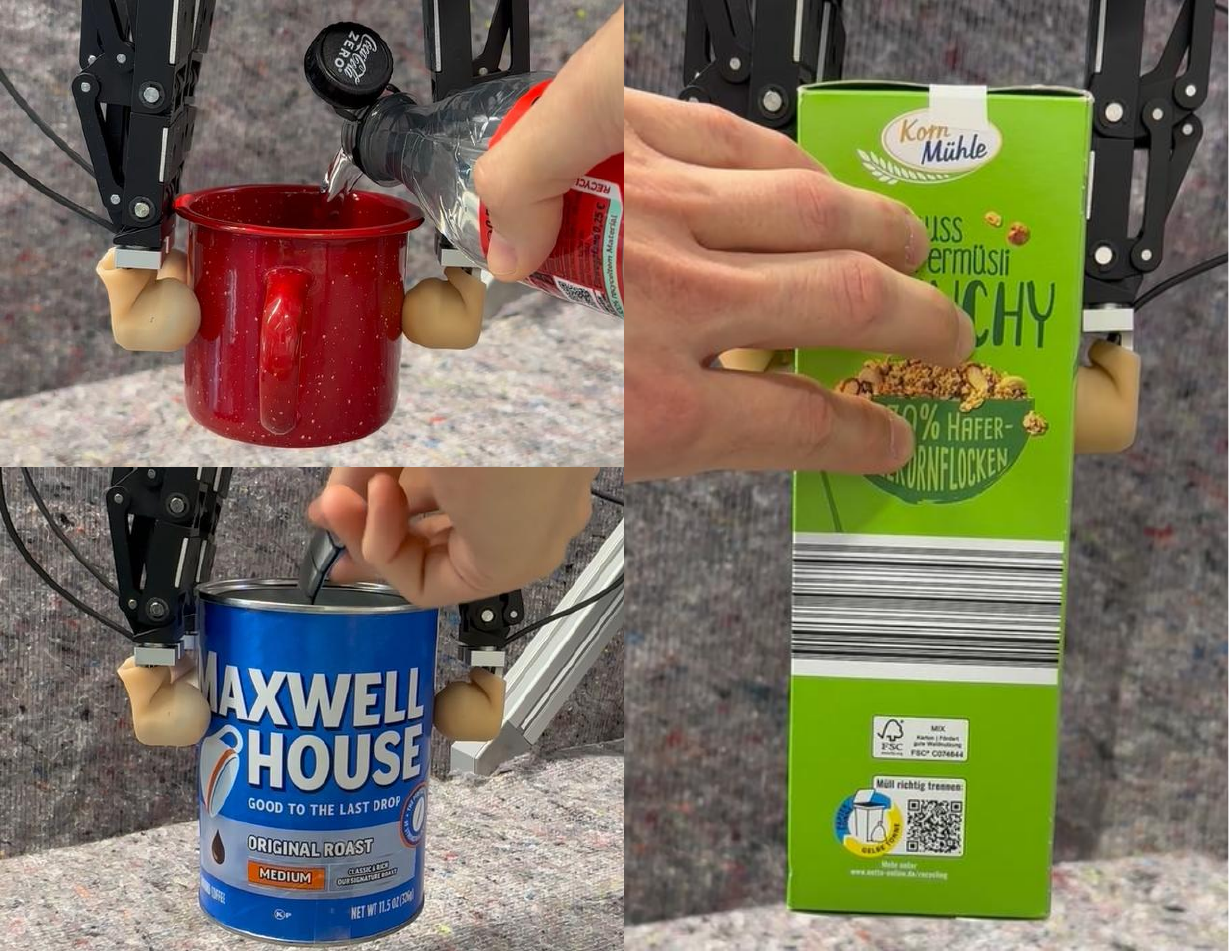}}
    \caption{\textbf{Disturbance events.} Images showing which disturbance events are used for the experiments to test if the controller is able to react to disturbances.}
    \label{fig:disturbanceExperiments}
\end{figure}

\begin{figure}
    \centerline{\includegraphics[width=\columnwidth]{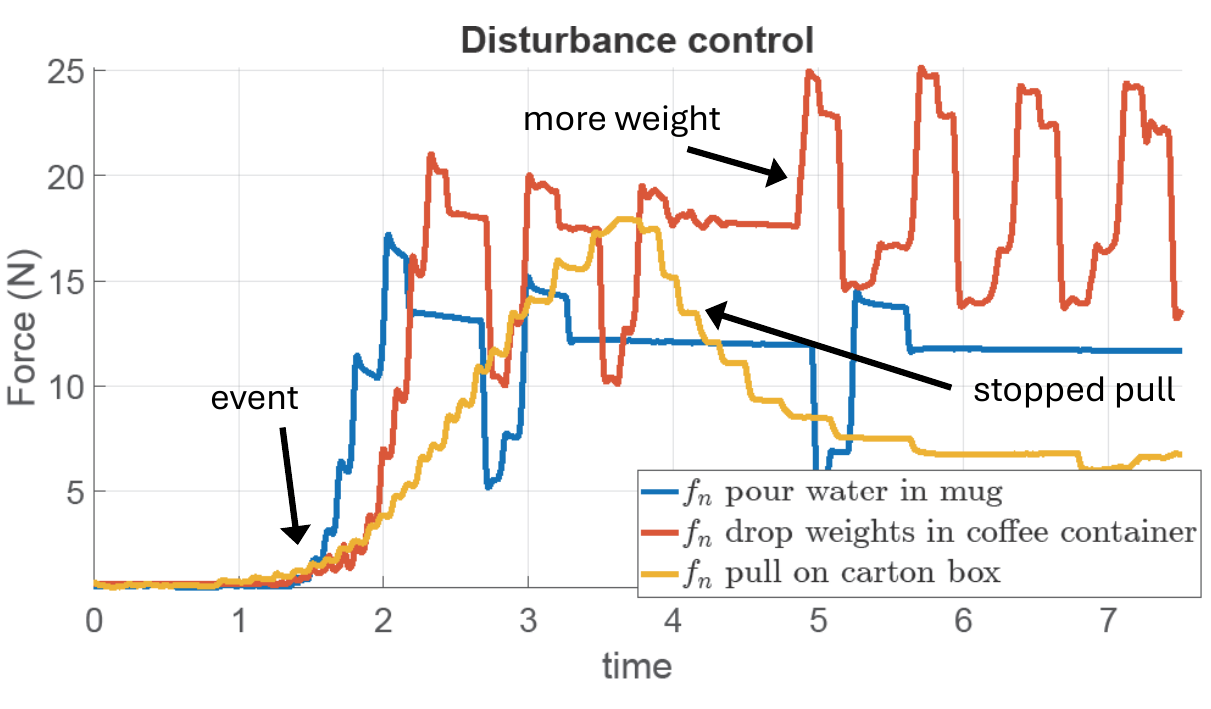}}
    \caption{\textbf{Disturbance during grasping.} Measured normal forces when disturbances act on initially safely grasped objects.}
    \label{fig:disturbancePlot}
\end{figure}

In the last experiment, our adaptive force controller was tested against disturbances that occurred after a safe grasp was established. We tested again on deformable and stiff objects. \fref{fig:disturbanceExperiments} visualizes the events executed while the gripper held the object, including pouring water into a steel cup, dropping weights into a stiff coffee container, and human interaction with a carton box. The measured normal forces during this experiment are shown in \fref{fig:disturbancePlot}. Since the tested objects are very light, the force required to lift them initially is very low. At $\approx 1.5~s$ the event is executed and leads to a rise of the exerted force on the objects to prevent them from slipping. In all tests, the object was successfully prevented from dropping out of the grasp. No visual movement due to slippage was noticeable during the continuous pour of water into the cup or during the pulling of the paper box. Only when dropping the weights in the coffee container was slippage visible before the container was "caught" again. 

\section{Conclusion}
\label{sec:conclusion}

This paper addressed the problem of sensitive grasping using generic
grasping hardware. We presented a sensing and control framework
for safe robotic grasping and stable grasp maintenance. The system explores the minimal grasping forces required for a stable grasp of fragile and deformable objects and continuously analyzes contact relations with the object through our finger geometry design to maintain a stable grasp under varying load conditions.

A geometry-aware F/T-based contact estimation method is combined with an admittance controller to improve the grasp capabilities of a standard coarse-position-controlled gripper. The analysis of the contact point position on the finger, together with the normal and tangential force components, enables the computation of the desired grasping force without knowledge of the object’s physical properties. 

We achieved compliant, stable grasp acquisition with our approach that limits contact forces to prevent damage to the manipulated object. The experiments showed that the controller can safely and stably grasp objects with different shapes, stiffnesses, and friction coefficients. Furthermore, the system can respond to disturbances by adjusting the contact force to maintain a stable grasp.

\addtolength{\textheight}{-12cm}   %

\bibliographystyle{ieeetr}
\bibliography{IROS_2026}

\end{document}